\documentclass[10pt, journal,twoside]{IEEEtran} 

\usepackage{graphicx}
\usepackage{dblfloatfix}
\usepackage{mathptmx}
\usepackage{helvet}
\usepackage{courier}
\usepackage{textcomp}
\usepackage{amsmath}
\usepackage{balance}
\usepackage{amssymb}
\usepackage{cases}
\usepackage[font=footnotesize,labelfont=bf]{caption}
\usepackage{color}
\usepackage{multirow}
\usepackage{float}
\usepackage[linesnumbered,ruled]{algorithm2e}
\usepackage[marginal]{footmisc}

\makeatletter
  \newcommand\figcaption{\def\@captype{figure}\caption}
  \newcommand\tabcaption{\def\@captype{table}\caption}
\makeatother
\usepackage[numbers]{natbib}
\usepackage{epstopdf}
\usepackage{bm}
\usepackage{setspace}
\usepackage[backref]{hyperref} 
\hypersetup{
    hidelinks,
    colorlinks=false,
}
\usepackage{xcolor}
\usepackage{makecell}
\usepackage{booktabs}
\usepackage{tabularx}
\usepackage{colortbl}
\usepackage{subcaption}
\usepackage{float}
\usepackage{placeins}
\usepackage{cuted}

\begin{document}
\raggedbottom

\title{Temporal-Decay Shapley: A Time-Aware Data Valuation Framework for Time-Series Data}

\author{Chuwen Pang, Bing Mi, and Kongyang Chen
\IEEEcompsocitemizethanks{
\IEEEcompsocthanksitem Chuwen Pang and Kongyang Chen are with School of Artificial Intelligence, Guangzhou University, Guangzhou 510006, China.
\IEEEcompsocthanksitem Bing Mi is with School of Public Finance and Taxation, Guangdong University of Finance and Economics, Guangzhou 510320, China. 
}}

\IEEEtitleabstractindextext{
\begin{abstract}
With the rapid development of machine learning applications on time-series data, accurately assessing the value of training samples has become essential for data selection, noise detection, and model optimization. However, traditional data valuation methods usually assume that samples are independent and identically distributed, and thus ignore the time-varying nature of sample value in time-series data. This paper proposes an improved temporal Shapley data valuation method that enables accurate sample valuation for time-series data through a temporal decay mechanism and a multi-scale fusion strategy. Specifically, we propose three progressively enhanced temporal Shapley methods. Temporal-Decay Shapley (TDS) incorporates temporal information into Shapley value computation through exponential decay weights; the improved TDS adopts power exponential decay to better adapt to nonlinear temporal drift; and Multi-Scale Temporal-Decay Shapley (MS-TDS) constructs a multi-scale fusion mechanism that balances the value of short-term hotspot samples and long-term foundational samples through parallel multi-scale valuation and sample-level adaptive fusion. Experimental results show that the proposed methods generally outperform traditional methods in noise detection and high-value data identification tasks, with more evident advantages under most strongly temporal settings, thereby effectively improving the accuracy and robustness of data valuation.
\end{abstract}
\begin{IEEEkeywords}
Data valuation, Shapley value, time-series data, temporal decay, multi-scale fusion.
\end{IEEEkeywords}
}

\maketitle
\IEEEdisplaynontitleabstractindextext
\IEEEpeerreviewmaketitle

\section{Introduction}

In recent years, with the rapid development of artificial intelligence, machine learning models have achieved remarkable success in various fields such as computer vision, natural language processing, financial risk control, and medical diagnosis. Data-driven intelligent decision-making has become an important feature of modern society. However, during model training, different training samples often contribute differently to the final model performance. Such differences are reflected not only in the amount of information contained in the samples, but also in their timeliness and applicability. Accurately evaluating the value of training samples is of great importance for key tasks such as data selection, noise detection, and model optimization. It can help construct more compact and effective training sets, identify and remove harmful noisy samples, and guide the design of data augmentation strategies~\cite{ghorbani2019data,jia2019towards}.

In traditional machine learning frameworks, data are typically regarded as a static collection of independent and identically distributed samples, and each sample is assumed to have equal learning value. However, this assumption often fails to hold in real-world applications. In particular, in time-series data scenarios, the value of samples may change significantly over time. Recently generated data often better reflect the current data distribution and user behavior patterns, whereas historical data may gradually lose their reference value due to environmental changes, concept drift, and other factors~\cite{gama2014survey,webb2016characterizing}. Such differences in timeliness are especially prominent in real-time applications such as financial trading, network traffic analysis, and sensor monitoring.

The Shapley value, a classical concept in cooperative game theory~\cite{shapley1953value}, has become a core theoretical framework for data valuation because it rigorously satisfies four fundamental axioms: efficiency, symmetry, additivity, and dummy player. This theory was originally proposed by Lloyd Shapley in 1953. By calculating the marginal contribution of each participant, it determines the deserved payoff share of each participant and ensures the fairness and rationality of allocation. Current mainstream data valuation methods, such as Leave-One-Out (LOO) and Monte Carlo Shapley approximation methods~\cite{castro2009polynomial}, are all based on the core assumption that training samples are independent and identically distributed (IID). They construct utility functions based on prediction probabilities or classification accuracy to statically evaluate sample value.

However, in practical applications, many datasets exhibit clear temporal correlations and concept drift, which directly challenge the theoretical foundation of traditional methods. Data timeliness is an important dimension of data quality and has a significant impact on sample value. In time-series data scenarios, the value of a sample is often closely related to its generation time. Recent samples are more consistent with the current data distribution and model learning objectives, and the patterns they contain are more beneficial for improving model performance. By contrast, historical samples may become less informative due to environmental evolution and pattern changes. If all samples are valued without distinction, biases such as underestimating highly timely samples and overestimating outdated samples may occur~\cite{kifer2004detecting,tsymbal2004problem}.

Traditional Shapley value methods face three main limitations when applied to time-series data. First, temporal information is lost, because traditional methods treat samples with different timestamps as equivalent individuals and fail to capture value differences along the temporal dimension. Second, robustness is insufficient. In concept drift scenarios, traditional methods may mistakenly assign low values to temporally important samples. Third, adaptability is limited, because the fixed weighting mechanism of traditional methods cannot accommodate the temporal characteristics of different datasets~\cite{kwon2022beta,xia2024pshapley}.

To address the above problems, this paper proposes an improved temporal Shapley data valuation method. The core idea is to use a temporal decay mechanism and a multi-scale fusion strategy to enable accurate valuation of samples in time-series data, while reducing valuation bias caused by the absence of temporal information. By treating the temporal dimension as a key regulating factor of sample value, the proposed method effectively couples ``information value'' and ``temporal value'', providing a new solution for data valuation in temporally non-stationary scenarios.

The main contributions of this paper are summarized as follows:

\begin{itemize}
    \item We explicitly incorporate the temporal dimension into the Shapley value computation framework. Through a temporal decay mechanism and a multi-scale fusion strategy, we provide a new theoretical paradigm for valuing temporally non-stationary data. Three progressively enhanced temporal Shapley methods are proposed, ranging from basic temporal decay to multi-scale adaptive fusion, forming a complete theoretical framework for time-series data valuation.

    \item We design three progressively enhanced temporal Shapley methods, establishing a complete technical system from basic to advanced valuation. The TDS method effectively incorporates temporal information through exponential decay weights; the improved TDS adopts power exponential decay to better adapt to nonlinear temporal changes; and MS-TDS achieves a balance between the values of short-term hotspot samples and long-term foundational samples through multi-scale parallel computation and adaptive fusion.

    \item The proposed methods exhibit strong engineering applicability and can effectively improve data cleaning efficiency and model robustness, providing valuable tools for practical applications. Experimental results on multiple heterogeneous datasets show that the proposed methods generally outperform traditional methods in noise detection and high-value data identification tasks under most settings.
\end{itemize}

Organization. The rest of this paper is organized as follows. Section~2 presents the research motivation and background of the main contributions. Section~3 provides the problem formulation and theoretical modeling. Section~4 details the improved temporal Shapley framework and algorithm design. Section~5 discusses the experimental design and key results. Section~6 reviews related research progress. Section~7 concludes the paper and outlines future directions.

\section{Motivation}

\subsection{Time-Series Data Valuation}

Data valuation aims to quantify the marginal contribution of each training sample to model utility, and is commonly used in data selection, noise detection, training set reduction, and cost-constrained priority annotation. For a given learning task, a ``valuable sample'' usually satisfies two conditions. First, it provides additional discriminative information and improves the utility function $U(\cdot)$ on the validation set. Second, it remains effective under the current distribution of the target application and helps the model adapt to the latest patterns in real-world environments.

Under the static IID setting, sample value can be regarded as a relatively stable quantity, and classical Shapley valuation provides a reasonable characterization in the sense of fairly allocating marginal contributions. However, in time-series scenarios, the data generation mechanism often changes over time due to concept drift, periodicity, sudden events, and other factors. As a result, sample value exhibits significant temporal heterogeneity. Recent samples are closer to the current distribution and are usually more critical to short-term performance, whereas historical samples may contain long-term regularities or extreme patterns that still contribute to robustness and generalization. In other words, sample value is no longer determined solely by information contribution, but is jointly modulated by temporal position and temporal scale.

Therefore, data valuation for time-series data needs to answer a more specific question: when the model objective is to serve prediction at the current time or over a future period, which historical segments should be retained and assigned higher weights, and which segments should be down-weighted or even removed, so as to achieve a better balance among accuracy, robustness, and computational cost?

\subsection{Why Time-Aware and Multi-Scale}

The core assumption of traditional Shapley value methods in data valuation is IID. These methods quantify value only through marginal contributions, but ignore the fact that information changes dynamically over time in time-series scenarios. In tasks involving concept drift or sensitivity to a valid time window, treating samples with different timestamps equally may lead to biases such as underestimating highly timely samples and overestimating outdated samples, thereby weakening the adaptability of valuation results to dynamic scenarios~\cite{kwon2022beta,xia2024pshapley}. In practice, this bias often manifests as the model being dominated by a large number of outdated historical samples, which weakens its learning of the latest distribution and eventually causes performance degradation during deployment or rolling prediction.

From the perspective of method design, time-series data valuation faces at least three key challenges:

\begin{itemize}
    \item \textbf{Temporal correlation and dependency structure.}
    Time-series samples often exhibit strong correlations and continuity. The value of certain samples may not come from their isolated point-wise contribution, but from the learnable patterns jointly formed with their neighboring segments. If valuation relies only on marginal contributions without temporal distinction, the overall effect of key temporal segments may be underestimated.

    \item \textbf{Concept drift and temporal decay.}
    When the distribution evolves over time, older samples are more likely to mismatch the current task. Therefore, a mechanism that decreases with the time gap $\Delta t$ is needed, so that valuation can explicitly reflect timeliness.

    \item \textbf{Coexistence of short-term hotspots and long-term regularities.}
    Relying on a single temporal scale may cause a biased valuation. Overemphasizing recent samples may lose long-term regularities and extreme patterns, whereas overemphasizing historical samples may sacrifice adaptability to the current distribution. In practical applications, short-term changes determine immediate performance, while long-term regularities affect robustness and generalization. Both aspects should be characterized simultaneously.
\end{itemize}

In addition, time-series data often contain multi-scale information such as short-term hotspots and long-term regularities. A single decay form is difficult to accommodate both aspects: short-term scales better reflect the current distribution, while long-term scales better support generalization ability~\cite{mallat1989theory}. Therefore, a valuation method is needed that can jointly model temporal timeliness and multi-scale characteristics within a unified framework.

Motivated by the above observations, this paper adopts a combined strategy of temporal decay and multi-scale modeling. First, temporal decay is used to explicitly model the timeliness of samples, allowing valuation to decrease controllably with $\Delta t$. Furthermore, multi-scale parallel valuation and sample-level adaptive fusion are introduced, so that value information from different temporal scales can be integrated within a unified framework. This enables the method to simultaneously account for short-term adaptability and long-term robustness. A direct and testable expectation is that, in noise detection tasks, the ranking of low-value samples should better match noise patterns that disrupt temporal regularities; in data selection or removal tasks, model utility should decline more rapidly when high-value samples are removed first, thereby demonstrating the effectiveness of the valuation ranking.

\section{Problem Formulation}

To address the limitations of traditional Shapley value methods in time-series data, this paper proposes three progressively enhanced methods, forming a complete technical framework from basic to advanced valuation. While preserving the theoretical rigor of the Shapley value, these three methods incorporate the temporal dimension into the data valuation process through different technical routes, thereby enabling accurate quantification of the value of time-series data.

\subsection{Notations and Data}

Let the training dataset be
\begin{equation}
D = \{(x_i, y_i, t_i)\}_{i=1}^{N},
\end{equation}
where $x_i \in \mathbb{R}^{d}$ denotes the feature vector, $y_i \in \mathcal{Y}$ denotes the label, and $t_i$ denotes the timestamp. This paper focuses on the supervised learning setting, with classification tasks as the main example. Let $\mathcal{A}(\cdot)$ denote the learning algorithm, which takes a training subset $S \subseteq D$ as input and outputs a model $f_S = \mathcal{A}(S)$.

Let the validation set be
\begin{equation}
D_{\mathrm{val}} = \{(x^{\mathrm{val}}_j, y^{\mathrm{val}}_j)\}_{j=1}^{N_{\mathrm{val}}},
\end{equation}
which is used to measure model utility. The utility function $U(S)$ represents the evaluation result on $D_{\mathrm{val}}$ after training $f_S$ on the training subset $S$, such as accuracy, AUC, or negative log-likelihood (NLL). For simplicity, this paper assumes that a larger value of $U(\cdot)$ indicates better model performance.

To highlight the temporal factor, we define a reference time $t_{\mathrm{ref}}$, which is usually set as the latest timestamp among the data visible during training, and define the time gap as
\begin{equation}
\Delta t_i = t_{\mathrm{ref}} - t_i.
\end{equation}
When the data distribution evolves over time, $\Delta t_i$ characterizes the relative freshness or staleness of a sample, and is therefore an important variable affecting its temporal value.

\subsection{Classical Data Shapley}

In the traditional data valuation framework, the value of a sample is computed using the Shapley value:
\begin{equation}
\phi_i =
\sum_{S \subseteq D \setminus \{x_i\}}
\frac{|S|!(|D|-|S|-1)!}{|D|!}
\cdot
\left[ U(S \cup \{x_i\}) - U(S) \right],
\end{equation}
where $U(S)$ denotes the utility of the model trained on subset $S$ and evaluated on an independent validation set, such as the average prediction probability or classification accuracy.

\subsection{Time-Aware Valuation Objective}

However, in time-series data scenarios, the value of a sample depends not only on its information contribution, but also on temporal factors. Intuitively, when concept drift or valid-window effects exist, samples closer to $t_{\mathrm{ref}}$ are often more representative of the current distribution. Meanwhile, some historical samples may still contain regularities that are important for long-term generalization. Therefore, it is necessary to introduce a controllable temporal modulation mechanism into the Shapley framework for fairly allocating marginal contributions.

\textbf{Problem definition.}
Given a timestamped training set $D$, a validation set $D_{\mathrm{val}}$, and a utility function $U(\cdot)$, the goal is to learn a sample-level valuation vector $\phi^{\mathrm{TDS}} \in \mathbb{R}^{N}$ that simultaneously characterizes: (1) the marginal contribution of a sample to model utility, namely its information value; and (2) the timeliness of a sample relative to $t_{\mathrm{ref}}$, namely its temporal value. This can be formally expressed as
\begin{equation}
\phi^{\mathrm{TDS}}_i =
f\left(\phi^{\mathrm{Shapley}}_i, \Delta t_i, \lambda\right),
\end{equation}
where $\phi^{\mathrm{Shapley}}_i$ denotes the traditional Shapley value, and $\lambda$ is the temporal decay parameter.

\begin{figure*}[t]
    \centering
    \includegraphics[width=1\textwidth]{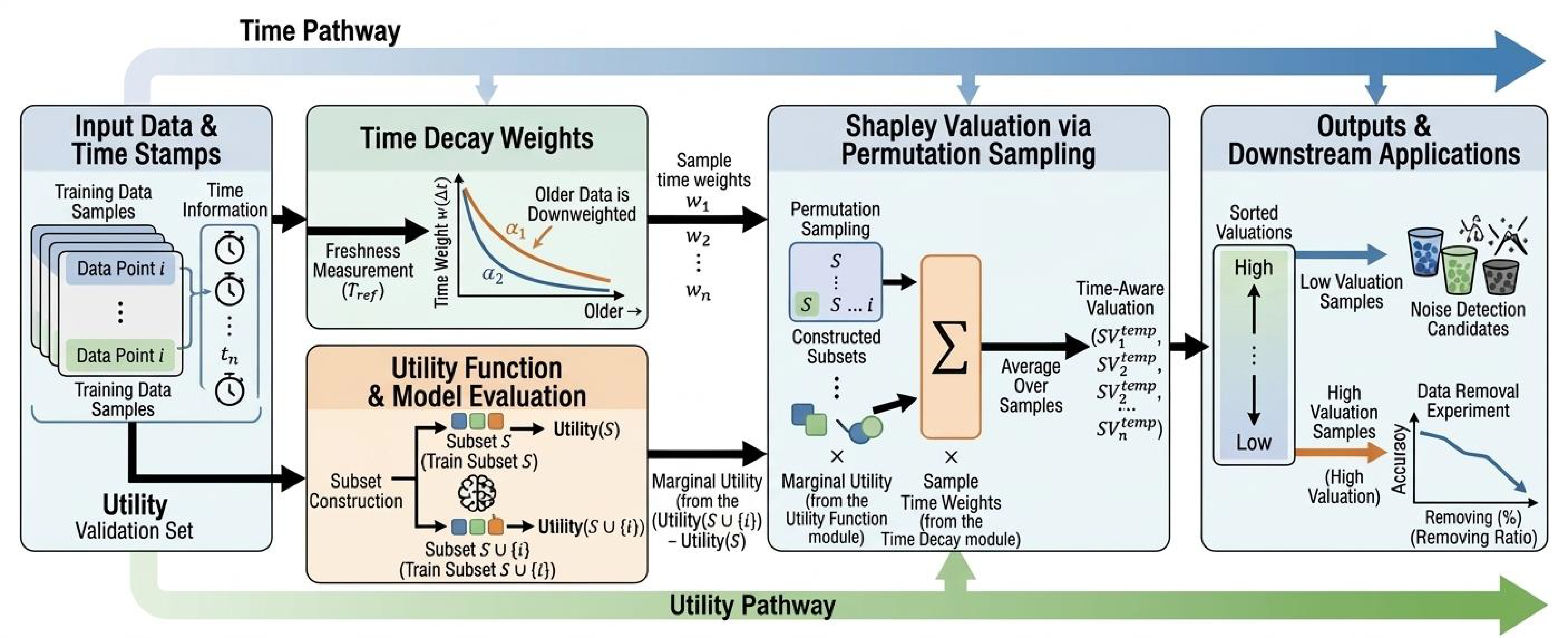}
    \caption{Overview of the temporal Shapley valuation framework. The upper temporal pathway maps sample freshness or staleness to temporal weights, while the lower utility pathway provides marginal utilities. The two pathways are combined in the Shapley approximation process based on permutation sampling, producing sample valuations for downstream tasks such as noise detection and data removal.}
    \label{fig:framework}
\end{figure*}

In the model construction, the methods proposed in the following sections explicitly define a temporal weight $w_i(\Delta t_i)$ and use it to weight marginal contributions, thereby obtaining time-aware valuation results. This enables the valuation to exhibit controllable decay or nonlinear variation with respect to $\Delta t$. The overall framework is illustrated in Fig.~\ref{fig:framework}: the temporal pathway maps the freshness or staleness of samples to temporal weights, while the utility pathway provides marginal utilities. The two pathways are combined in the approximation process based on permutation sampling, producing sample valuation results that can be further used for downstream tasks such as noise detection and data selection/removal.

\section{Methodology}

To address the limitations of traditional Shapley value methods in time-series data, this paper proposes three progressively enhanced methods, forming a complete technical framework from basic to advanced valuation. While preserving the theoretical rigor of the Shapley value, these three methods incorporate the temporal dimension into the data valuation process through different technical routes, thereby enabling accurate quantification of the value of time-series data.

\subsection{Temporal-Decay Shapley (TDS)}

\subsubsection{Core Idea}

The core idea of TDS is to treat the temporal dimension as a key regulating factor of sample value. By introducing a temporal decay mechanism, the computation of the Shapley value jointly couples the information contribution of a sample and its temporal timeliness. In this way, TDS preserves the fairness and rigor of the Shapley value under the cooperative game-theoretic framework, while accurately characterizing the dynamic value of time-series data.

\subsubsection{Mathematical Definition}

Based on the time gap $\Delta t_i$, an exponential decay weight $w_i$ is introduced to characterize the regulating effect of time on sample value. It is defined as
\begin{equation}
w_i = \exp(-\lambda \Delta t_i),
\end{equation}
where $\lambda$ is the decay coefficient that controls the overall decay intensity. The baseline TDS adopts the above exponential form. When stronger nonlinear temporal changes need to be modeled, the power exponential decay introduced in Section~\ref{subsec:improved_tds} can be used as an extension.

Let the reference time be $t_{\mathrm{ref}}$, and let the collection time of sample $i$ be $t_i$. The time gap is defined as
\begin{equation}
\Delta t_i = \max(0, t_{\mathrm{ref}} - t_i).
\end{equation}
To eliminate the influence of measurement units, $\Delta t_i$ can be further normalized by a selected characteristic time scale $T_{\mathrm{ref}}$ before being substituted into the weighting function.

By incorporating temporal decay weights, the TDS value is defined as the expected weighted marginal contribution of a sample over all possible subset permutations:
\begin{equation}
\phi^{\mathrm{TDS}}_i =
\sum_{S \subseteq D \setminus \{x_i\}}
\frac{|S|!(|D|-|S|-1)!}{|D|!}
\cdot w_i \cdot
\left(
U(S \cup \{x_i\}) - U(S)
\right).
\end{equation}

\subsubsection{Algorithm Implementation}

Since the exact computation of the Shapley value has a computational complexity of $O(2^N)$, permutation-sampling-based approximation algorithms are commonly adopted in practical applications. The approximation procedure of TDS is shown in Algorithm~\ref{alg:tds}. The time complexity of this algorithm is $O(M \cdot N \cdot C)$, where $C$ denotes the cost of a single model training and evaluation process. By adjusting the number of samples $M$, a balance can be achieved between computational accuracy and efficiency.

\begin{algorithm}[t]
\caption{Approximate Computation of TDS}
\label{alg:tds}
\KwIn{Training data $(X_{\mathrm{train}}, y_{\mathrm{train}})$; validation data $(X_{\mathrm{val}}, y_{\mathrm{val}})$; classifier $clf$; timestamps; decay rate $\lambda$; number of samples $M$}
\KwOut{TDS valuation vector $\phi$}

Initialize $\phi = 0$\;
Compute time gaps: $\Delta t = \mathrm{current\_time} - \mathrm{timestamps}$\;
Compute decay weights: $w_i = \exp(-\lambda \cdot \Delta t_i)$\;

\For{$m \leftarrow 1$ \KwTo $M$}{
    Generate a random permutation $\pi = \mathrm{random\_permutation}(N)$\;
    Initialize subset $S = \emptyset$ and previous performance $prev\_perf = \mathrm{None}$\;
    
    \For{$i \in \pi$}{
        $S \leftarrow S \cup \{x_i\}$\;
        Train classifier: $clf.fit(X_S, y_S)$\;
        Compute current performance:
        $curr\_perf = accuracy(clf.predict(X_{\mathrm{val}}), y_{\mathrm{val}})$\;
        
        \If{$prev\_perf \neq \mathrm{None}$}{
            $\phi[i] \leftarrow \phi[i] + w_i \cdot (curr\_perf - prev\_perf)$\;
        }
        $prev\_perf \leftarrow curr\_perf$\;
    }
}
\Return{$\phi/M$}\;
\end{algorithm}

\subsection{Improved TDS}
\label{subsec:improved_tds}

\subsubsection{Method Motivation}

Although the exponential decay function used in the basic TDS method can adapt to scenarios with linear temporal drift, it has limitations in the following two types of complex time-series scenarios:

\begin{enumerate}
    \item \textbf{Nonlinear temporal drift.}
    In some tasks, the decay of sample value exhibits nonlinear characteristics. The uniformly decreasing nature of exponential decay cannot accurately fit such nonlinear trends.

    \item \textbf{Valid-window-sensitive scenarios.}
    In some tasks, data have a clear valid time window. However, the gradual decay of the exponential function may still assign relatively high weights to samples outside the valid window.
\end{enumerate}

\subsubsection{Design of the Power Decay Function}

In this paper, the improved TDS adopts the power exponential decay mode, which is defined as
\begin{equation}
w_i = \exp(-\lambda \Delta t_i^p),
\end{equation}
where $p$ is a nonlinear regulation parameter. When $p < 1$, the decay rate slows down as $\Delta t_i$ increases; when $p > 1$, the decay rate accelerates as $\Delta t_i$ increases; and when $p = 1$, the function degenerates into standard exponential decay.

\subsubsection{Algorithm Implementation}

The core algorithm of the improved TDS is built upon the basic TDS algorithm. It introduces power exponential decay weights and controls the decay rate and degree of nonlinearity through the parameters $(\lambda, p)$. In addition, a positive scaling coefficient $\alpha > 0$ can be introduced, and the valuation can be uniformly written as $\phi = \alpha \phi^{\mathrm{TDS}}$. This coefficient can be used to match numerical scales under different utility metrics or serve as an interface for linear combination with other additive valuation terms.

\begin{algorithm}[t]
\caption{Improved TDS Algorithm}
\label{alg:improved_tds}
\KwIn{Training data $(X_{\mathrm{train}}, y_{\mathrm{train}})$; validation data $(X_{\mathrm{val}}, y_{\mathrm{val}})$; classifier $clf$; timestamps; power decay parameters $(\lambda, p)$; scaling coefficient $\alpha$; number of samples $M$}
\KwOut{Improved TDS valuation vector $\phi$}

Initialize $\phi = 0$\;
Compute time gaps: $\Delta t = \mathrm{current\_time} - \mathrm{timestamps}$\;
Compute power exponential decay weights: $w_i = \exp(-\lambda \cdot (\Delta t_i)^p)$\;

\For{$m \leftarrow 1$ \KwTo $M$}{
    Generate a random permutation $\pi = \mathrm{random\_permutation}(N)$\;
    Initialize subset $S = \emptyset$ and previous performance $prev\_perf = \mathrm{None}$\;
    
    \For{$i \in \pi$}{
        $S \leftarrow S \cup \{x_i\}$\;
        Train classifier: $clf.fit(X_S, y_S)$\;
        Compute current performance:
        $curr\_perf = accuracy(clf.predict(X_{\mathrm{val}}), y_{\mathrm{val}})$\;
        
        \If{$prev\_perf \neq \mathrm{None}$}{
            $\phi[i] \leftarrow \phi[i] + w_i \cdot (curr\_perf - prev\_perf)$\;
        }
        $prev\_perf \leftarrow curr\_perf$\;
    }
}
\Return{$\alpha \cdot (\phi/M)$}\;
\end{algorithm}

\subsection{Multi-Scale Temporal-Decay Shapley (MS-TDS)}

\subsubsection{Method Motivation}

Both the basic TDS and the improved TDS follow a single-scale temporal decay logic, which has clear limitations for multi-scale time-series data. In real-world time-series tasks, samples often contain both short-term hotspot information and long-term foundational information. Short-term information reflects immediate patterns in the data and is crucial for adapting the model to the current distribution, whereas long-term information reflects fundamental regularities and is indispensable for maintaining model generalization ability.

\subsubsection{Multi-Scale Computation}

The multi-scale design of MS-TDS follows the principle of prioritizing physical interpretability. The scale set is defined as $\{\tau_1, \tau_2, \tau_3\}$. To ensure that the decay rate of each scale matches its temporal granularity, each scale $\tau_k$ is transformed into a corresponding decay coefficient $\lambda_k$:
\begin{equation}
\lambda_k = \frac{\lambda}{\tau_k},
\end{equation}
where $\lambda$ is a predefined baseline decay intensity, which has the same meaning as in the single-scale TDS, and $\tau_k$ denotes the temporal granularity of the $k$-th scale. A larger $\tau_k$ corresponds to a smaller $\lambda_k$, resulting in smoother weight decay and thus placing greater emphasis on sample contributions over longer periods.

\subsubsection{Multi-Scale Parallel Valuation}

For each scale $k$, the temporal weight $w_i^{(k)}$ is calculated using the decay coefficient $\lambda_k$ of that scale. It is then combined with Shapley marginal contributions to obtain the valuation of sample $i$ at scale $k$:
\begin{equation}
\phi_i^{(k)} =
\sum_{S \subseteq D \setminus \{x_i\}}
\frac{|S|!(|D|-|S|-1)!}{|D|!}
\cdot w_i^{(k)}
\cdot
\left[
U(S \cup \{x_i\}) - U(S)
\right].
\end{equation}

\subsubsection{Sample-Level Adaptive Fusion}

This paper adopts sample-level adaptive fusion based on cross-scale summation and inverse-variance rescaling. The dispersion of multi-scale valuations is used to characterize uncertainty, and a stability-related coefficient is applied to the cross-scale aggregated result. Let the valuation vector of the $i$-th sample across $K$ scales be
\begin{equation}
\Phi_i = [\phi_i^{(1)}, \ldots, \phi_i^{(K)}].
\end{equation}
The fusion process is defined as follows.

\begin{enumerate}
    \item Variance statistic:
    \begin{equation}
    \sigma_i^2 = \mathrm{Var}(\Phi_i).
    \end{equation}

    \item Sample-level stability coefficient:
    \begin{equation}
    a_i = \frac{1}{\sigma_i^2 + \epsilon}.
    \end{equation}

    \item Final valuation:
    \begin{equation}
    \phi_i^{\mathrm{MS}} =
    a_i \sum_{k=1}^{K} \phi_i^{(k)}.
    \end{equation}
\end{enumerate}

This formulation is equivalent to first linearly aggregating the multi-scale valuations and then applying an overall rescaling based on the sample-level stability coefficient. Samples with higher cross-scale consistency, namely smaller variance, obtain larger stability coefficients $a_i$ and are therefore more strongly emphasized in the final valuation. In contrast, samples with larger cross-scale disagreement, namely larger variance, are relatively suppressed to reduce valuation noise caused by scale uncertainty. In implementation, $\epsilon$ is used to avoid numerical amplification when the variance is too small. The number of scales $K$ and the selection of $\{\tau_k\}$ should be consistent with the sampling interval and the major identifiable periodic scales in the data, and should be kept consistent across datasets for fair comparison.

\subsubsection{Algorithm Implementation}

In summary, TDS, improved TDS, and MS-TDS share the same implementation backbone in terms of permutation sampling and marginal utility evaluation. Their main differences lie in the construction of temporal weights and whether cross-scale fusion is performed. In the following experiments, under the same training-validation split and utility definition, these three methods and the baselines are compared in a unified manner for noise detection and high-value data removal. The experimental section first describes the datasets, baselines, and evaluation metrics, and then presents the main tables and curves for noise detection and high-value removal, so as to align the methodology described in this section with the empirical results reported later.

Accordingly, MS-TDS first calls Algorithm~\ref{alg:improved_tds} at each scale $k$ to obtain $\phi^{(k)}$, and then estimates cross-scale dispersion at the sample level and performs inverse-variance rescaling. The pseudo-code is shown in Algorithm~\ref{alg:ms_tds}.

\begin{algorithm}[t]
\caption{MS-TDS Multi-Scale Adaptive Fusion Algorithm}
\label{alg:ms_tds}
\KwIn{Training data $(X_{\mathrm{train}}, y_{\mathrm{train}})$; validation data $(X_{\mathrm{val}}, y_{\mathrm{val}})$; classifier $clf$; timestamps; scale set $\{\tau_k\}$; parameters $(\lambda, \epsilon)$}
\KwOut{MS-TDS valuation vector $\phi^{\mathrm{MS}}$}

Initialize $K \leftarrow |\{\tau_k\}|$, $N \leftarrow |X_{\mathrm{train}}|$\;

\For{$k \leftarrow 1$ \KwTo $K$}{
    Compute scale-specific decay rate: $\lambda_k \leftarrow \lambda/\tau_k$\;
    Run Algorithm~\ref{alg:improved_tds} to obtain scale-specific valuation $\phi^{(k)}$\;
}

\For{$i \leftarrow 1$ \KwTo $N$}{
    Compute mean:
    $\bar{\phi}_i \leftarrow \frac{1}{K}\sum_{k=1}^{K}\phi_i^{(k)}$\;
    
    Compute variance:
    $\sigma_i^2 \leftarrow \frac{1}{K}\sum_{k=1}^{K}(\phi_i^{(k)}-\bar{\phi}_i)^2$\;
    
    Compute stability coefficient:
    $a_i \leftarrow \frac{1}{\sigma_i^2+\epsilon}$\;
    
    Compute fused valuation:
    $\phi_i^{\mathrm{MS}} \leftarrow a_i \sum_{k=1}^{K}\phi_i^{(k)}$\;
}
\Return{$\phi^{\mathrm{MS}}$}\;
\end{algorithm}

\section{Experiments}

\subsection{Experimental Setup}

\subsubsection{Datasets}

Table~\ref{tab:datasets} summarizes the basic statistics and temporal property labels of the four experimental subsets. The task meanings and feature profiles of each dataset are described as follows.

\textbf{Covertype dataset.}
This dataset corresponds to a classification task related to forest cover types. It contains a relatively large number of features, with 54 dimensions, and exhibits weak temporal characteristics, mainly reflected in collection batches or weak drift.

\textbf{Wind dataset.}
This dataset corresponds to a binary classification task related to wind power or meteorological data. It exhibits clear periodic changes and is therefore regarded as strongly time-series data, with 14-dimensional features.

\textbf{Electricity dataset.}
This dataset corresponds to a binary classification task related to power load. It contains dynamic characteristics such as intra-day and weekly periodic patterns, and is regarded as strongly time-series data, with 8-dimensional features.

\textbf{Traffic dataset.}
This dataset corresponds to a binary classification task for traffic congestion prediction. It contains features such as traffic flow, speed, accidents, hour, and weekday. The time span is relatively long, about 62.46 days, and the dataset exhibits strong temporal characteristics.

To ensure cross-dataset comparability, all experiments use subsets of the same scale and preserve the original temporal order. The time span is calculated as the difference between the maximum and minimum timestamps in each subset, consistent with Table~\ref{tab:datasets}. Unless otherwise specified, the data splitting strategy, training process, and utility function are kept identical across different methods, so as to avoid implementation differences interfering with the comparison of valuation methods themselves.

\begin{table*}[t]
\centering
\caption{Summary of time-series datasets used in the experiments.}
\label{tab:datasets}
\begin{tabular}{lcccc}
\toprule
Dataset & Total samples & Features & Time span (days) & Temporal property \\
\midrule
covertype & 1500 & 54 & 0.01 & Weakly time-series (batch-wise drift) \\
wind & 1500 & 14 & 0.01 & Strongly time-series (meteorological, periodicity) \\
electricity & 1500 & 8 & 0.01 & Strongly time-series (power load, periodicity) \\
traffic & 1500 & 5 & 62.46 & Strongly time-series (daily/weekly patterns) \\
\bottomrule
\end{tabular}
\end{table*}

\subsubsection{Baseline Methods}

We compare six types of methods, which can be divided into two groups: traditional Shapley variants and temporal Shapley methods. The former mainly optimize marginal contribution estimation under supervised predictive performance, whereas the latter explicitly model temporal correlation and timeliness.

\textbf{Traditional Shapley variants:}

\begin{itemize}
    \item \textbf{LOO (Leave-One-Out):}
    This method measures sample value by removing each sample individually and retraining the model, using the change in model loss or performance as the valuation criterion.

    \item \textbf{TMC-Shapley (Truncated Monte Carlo Shapley):}
    This method approximates Shapley values using truncated Monte Carlo permutation sampling~\cite{castro2009polynomial}.

    \item \textbf{Beta-Shapley:}
    This method introduces a Beta distribution to adjust the weights of sample subsets and reduces valuation noise by controlling the shape parameters of the distribution.
\end{itemize}

\textbf{Temporal Shapley methods proposed in this paper:}

\begin{itemize}
    \item \textbf{TDS (Temporal-Decay Shapley):}
    This is the basic temporal decay method, which adopts exponential decay weights.

    \item \textbf{Improved TDS:}
    This method extends TDS by adopting power exponential decay, thereby improving its adaptability to nonlinear temporal patterns.

    \item \textbf{MS-TDS (Multi-Scale Temporal-Decay Shapley):}
    This is an adaptive multi-scale method that performs multi-scale parallel valuation and sample-level adaptive fusion.
\end{itemize}

\subsubsection{Evaluation Metrics}

Considering two core applications of data valuation, namely identifying harmful samples and selecting key samples, this paper reports results from two perspectives: noise detectability and high-value ranking consistency. The corresponding metrics are described below.

\textbf{Noise data detection.}

\textbf{Noise identification AUC.}
After random noise, namely label flipping, is introduced into the training labels, samples whose valuation scores are lower than the median valuation score of all training samples are regarded as noise candidates. A binary classification task is then constructed, and the area under the ROC curve (AUC) is calculated.

\textbf{High-value data removal.}

The following three metrics are all based on weighted cumulative quantities derived from removal curves, consistent with the implementation. Specifically, performance differences between adjacent removal steps are first calculated and accumulated, and then weighted by $1/k$. For Brier score and cross-entropy, the sign is reversed so that larger values indicate more severe performance degradation after removing high-value samples.

\textbf{Weighted accuracy drop (WAD).}
Let the accuracy sequence obtained under different removal ratios be $\{a_k\}_{k=0}^{T}$, where $a_0$ denotes the performance before removal, and set $a_{T+1}=0$. Let $\Delta_k = a_{k-1} - a_k$ and $C_k = \sum_{j=1}^{k}\Delta_j$. Then,
\begin{equation}
\mathrm{WAD} =
\sum_{k=1}^{T+1}
\frac{1}{k} C_k .
\end{equation}
This definition characterizes the overall degradation of the removal curve rather than the performance difference at a single point.

\textbf{Weighted Brier score drop (WBD).}
For the Brier score sequence, a cumulative quantity $C_k^{(\mathrm{BS})}$ is constructed in the same way as above, and the sign is reversed:
\begin{equation}
\mathrm{WBD} =
-\sum_{k=1}^{T+1}
\frac{1}{k} C_k^{(\mathrm{BS})}.
\end{equation}
Here, $C_k^{(\mathrm{BS})}$ is obtained by accumulating adjacent differences in the Brier score sequence. A larger WBD indicates a more significant degradation in probabilistic prediction quality after high-value samples are removed.

\textbf{Weighted cross-entropy drop (WCD).}
Similarly, for the cross-entropy sequence,
\begin{equation}
\mathrm{WCD} =
-\sum_{k=1}^{T+1}
\frac{1}{k} C_k^{(\mathrm{CE})}.
\end{equation}
Here, $C_k^{(\mathrm{CE})}$ is obtained by accumulating adjacent differences in the cross-entropy sequence. A larger WCD indicates a more significant increase in uncertainty after high-value samples are removed.

\subsection{Experimental Results}

\subsubsection{Noise Data Detection}

Table~\ref{tab:noise_auc} reports the noise identification AUC on four datasets under LR and NB models. Overall, no single method consistently performs best across all settings. TMC-Shapley and Beta-Shapley achieve the highest AUC in several cases, while TDS, improved TDS, and MS-TDS remain competitive in most settings and are generally among the leading methods.

This phenomenon is consistent with the optimization objectives of different methods. Noise detection is essentially a supervised task, evaluating whether a sample harms predictive performance due to incorrect labels. Methods such as TMC-Shapley directly construct valuation scores around marginal predictive gains, and are therefore naturally aligned with this objective, often achieving better AUC results. In contrast, the core objective of TDS is to model temporal correlation and timeliness, such as recency and temporal consistency, rather than explicitly optimizing label correctness. Therefore, it may not always achieve the best performance under the definition of label-flipping noise.

Furthermore, the noise construction in this paper adopts random label flipping. Such synthetic noise is weakly correlated with the temporal structure itself, and therefore does not always align with the temporal weighting mechanism. In other words, although the TDS family may not always lead in noise AUC, this does not negate its value in time-varying data scenarios. When the task focuses on concept drift, non-stationary processes, or temporally correlated distribution shifts, temporal information provides a value dimension that is orthogonal to supervised noise detectability.

\begin{table*}[t]
\centering
\caption{Noise detection AUC of different valuation methods on four datasets with LR and NB models.}
\label{tab:noise_auc}
\begin{tabular}{lcccccccc}
\toprule
\multirow{2}{*}{Method} & \multicolumn{4}{c}{LR} & \multicolumn{4}{c}{NB} \\
\cmidrule(lr){2-5} \cmidrule(lr){6-9}
& covertype & wind & electricity & traffic & covertype & wind & electricity & traffic \\
\midrule
LOO & 0.525 & 0.467 & 0.534 & 0.442 & 0.430 & 0.476 & 0.455 & 0.491 \\
Beta-Shapley & 0.663 & 0.614 & 0.677 & \textbf{0.892} & 0.631 & 0.602 & 0.625 & 0.757 \\
TMC-Shapley & 0.711 & \textbf{0.714} & \textbf{0.806} & \textbf{0.892} & 0.680 & 0.712 & \textbf{0.741} & \textbf{0.804} \\
TDS & 0.721 & 0.705 & 0.753 & 0.864 & 0.663 & \textbf{0.734} & 0.728 & 0.773 \\
TDS-improved & \textbf{0.741} & 0.693 & 0.752 & 0.866 & 0.665 & 0.713 & 0.723 & 0.753 \\
MS-TDS & 0.738 & 0.699 & 0.763 & 0.878 & \textbf{0.689} & 0.716 & 0.721 & 0.766 \\
\bottomrule
\end{tabular}
\end{table*}

\subsubsection{High-Value Data Removal}

The WAD, WBD, and WCD results for high-value sample removal are reported in Table~\ref{tab:removal_lr} for LR and Table~\ref{tab:removal_nb} for NB. The following discussion first summarizes the rankings and curve patterns shown in the tables, and then presents the removal curves under LR and NB.

Tables~\ref{tab:removal_lr} and~\ref{tab:removal_nb} report three types of degradation metrics after high-value samples are removed. Compared with noise detection, the TDS family shows more stable advantages in this task. In particular, MS-TDS achieves the best results for almost all metrics under the LR setting, and also belongs to the leading group under the NB setting.

This result indicates that temporal information directly improves the quality of sample ranking. If a valuation method can more accurately identify samples that are most critical to the current distribution, then removing samples in descending order of valuation should lead to faster model performance degradation, corresponding to higher WAD, WBD, and WCD values. The advantages of MS-TDS across multiple datasets suggest that multi-scale fusion helps simultaneously capture short-term fluctuations and long-term structures, thereby improving the stability of high-value sample identification.

\begin{figure}[t]
    \centering
    \includegraphics[width=1\linewidth]{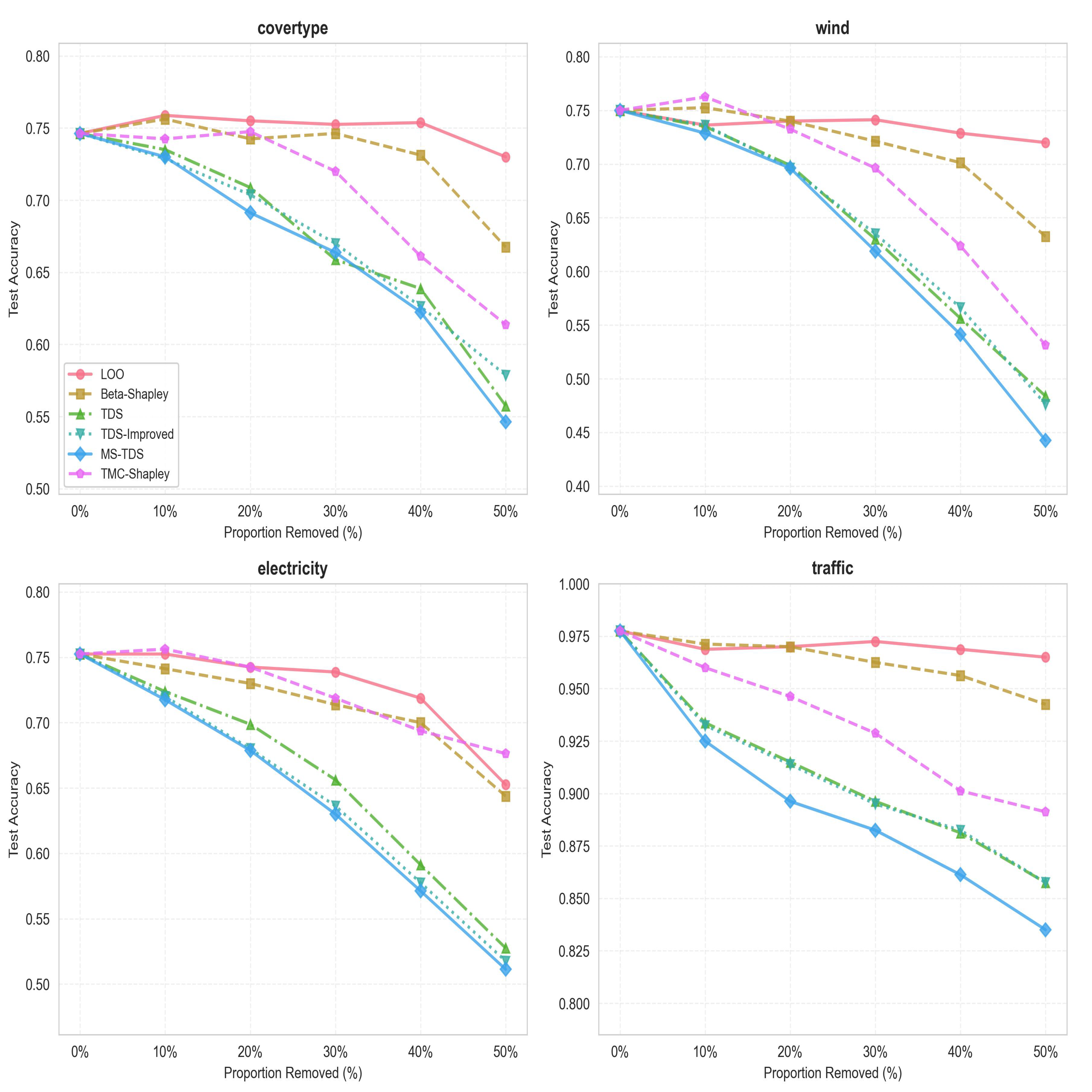}
    \caption{High-value data removal under the LR model, where the performance of different valuation methods changes as the removal ratio increases.}
    \label{fig:removal_lr}
\end{figure}

\begin{figure}[t]
    \centering
    \includegraphics[width=1\linewidth]{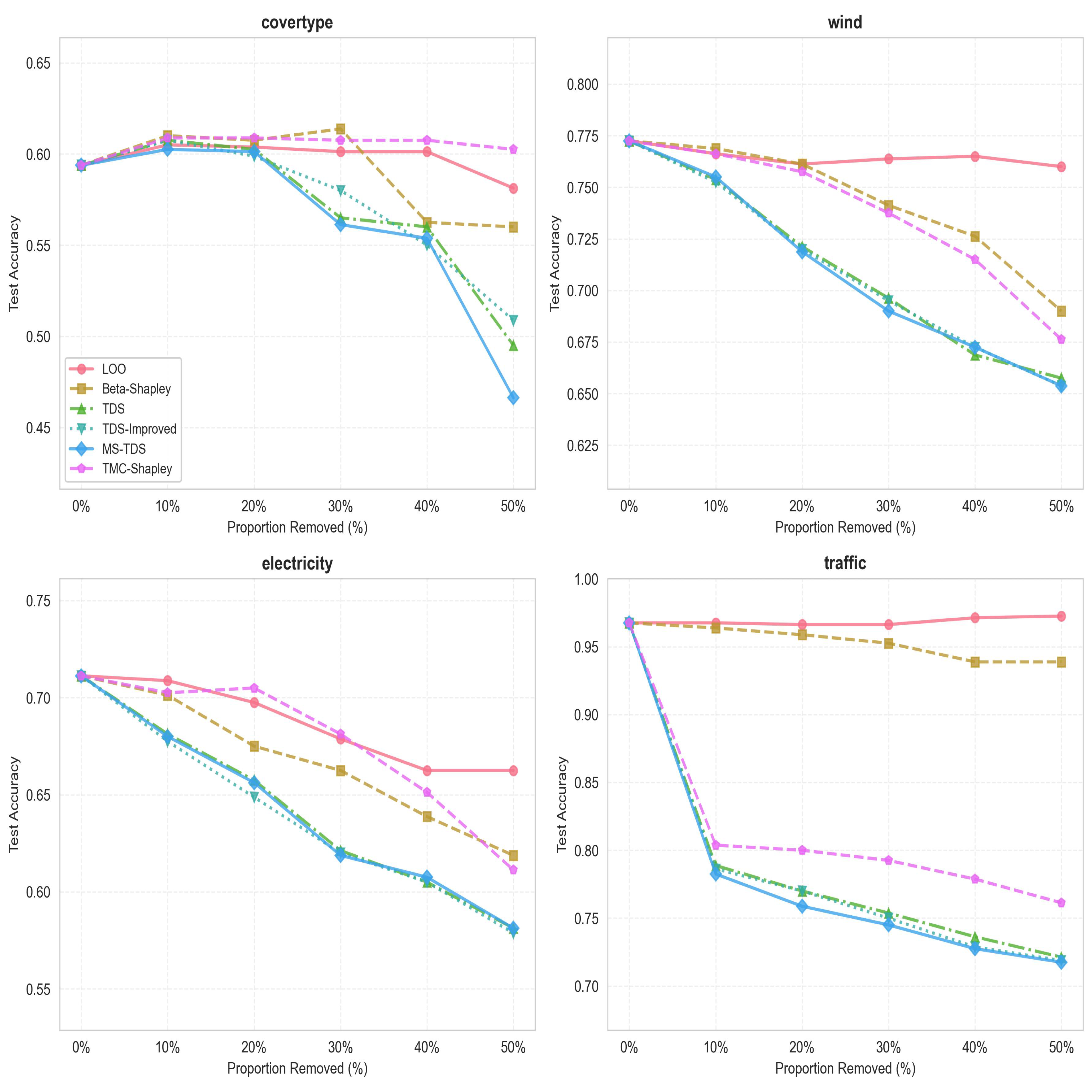}
    \caption{High-value data removal under the NB model, where the performance of different valuation methods changes as the removal ratio increases.}
    \label{fig:removal_nb}
\end{figure}

The removal curves in Fig.~\ref{fig:removal_lr} and Fig.~\ref{fig:removal_nb} are consistent with the conclusions from the tables. Time-sensitive methods usually trigger steeper performance degradation in the early and middle stages of removal, whereas some static baselines show slower degradation. This provides complementary evidence to the noise detection experiments: temporal valuation methods may not always be optimal for label noise identification, but they are more advantageous in data selection tasks driven by time-varying correlations.

\begin{table*}[t]
\centering
\caption{High-value data removal performance (LR): WAD, WBD, and WCD on four time-series datasets.}
\label{tab:removal_lr}
\resizebox{\textwidth}{!}{
\begin{tabular}{lcccccccccccc}
\toprule
\multirow{2}{*}{Method} 
& \multicolumn{3}{c}{covertype} 
& \multicolumn{3}{c}{wind} 
& \multicolumn{3}{c}{electricity} 
& \multicolumn{3}{c}{traffic} \\
\cmidrule(lr){2-4} \cmidrule(lr){5-7} \cmidrule(lr){8-10} \cmidrule(lr){11-13}
& WAD & WBD & WCD & WAD & WBD & WCD & WAD & WBD & WCD & WAD & WBD & WCD \\
\midrule
LOO & -0.022 & 0.052 & 0.667 & 0.128 & 0.101 & 1.179 & 0.126 & 0.120 & 1.792 & 0.165 & 0.024 & 0.067 \\
Beta-Shapley & 0.179 & 0.172 & 2.190 & 0.236 & 0.228 & 5.937 & 0.283 & 0.210 & 3.444 & 0.128 & 0.064 & 0.184 \\
TMC-Shapley & 0.235 & 0.215 & 2.129 & 0.304 & 0.296 & 4.332 & 0.228 & 0.169 & 2.354 & 0.246 & 0.115 & 0.329 \\
TDS & 0.328 & 0.305 & 2.425 & 0.412 & 0.421 & 7.018 & 0.430 & 0.352 & 4.453 & 0.384 & 0.182 & 0.543 \\
TDS-improved & 0.324 & 0.300 & 2.839 & 0.402 & 0.408 & 7.001 & 0.451 & 0.376 & 4.822 & 0.354 & 0.197 & 0.590 \\
MS-TDS & \textbf{0.367} & \textbf{0.338} & \textbf{3.463} & \textbf{0.431} & \textbf{0.443} & \textbf{7.840} & \textbf{0.466} & \textbf{0.404} & \textbf{6.110} & \textbf{0.414} & \textbf{0.225} & \textbf{0.696} \\
\bottomrule
\end{tabular}
}
\end{table*}

\begin{table*}[t]
\centering
\caption{High-value data removal performance (NB): WAD, WBD, and WCD on four time-series datasets.}
\label{tab:removal_nb}
\resizebox{\textwidth}{!}{
\begin{tabular}{lcccccccccccc}
\toprule
\multirow{2}{*}{Method} 
& \multicolumn{3}{c}{covertype} 
& \multicolumn{3}{c}{wind} 
& \multicolumn{3}{c}{electricity} 
& \multicolumn{3}{c}{traffic} \\
\cmidrule(lr){2-4} \cmidrule(lr){5-7} \cmidrule(lr){8-10} \cmidrule(lr){11-13}
& WAD & WBD & WCD & WAD & WBD & WCD & WAD & WBD & WCD & WAD & WBD & WCD \\
\midrule
LOO & 0.001 & 0.024 & 0.384 & 0.022 & 0.024 & 0.432 & 0.117 & 0.098 & 0.968 & 0.130 & 0.012 & 0.044 \\
Beta-Shapley & 0.184 & 0.147 & 0.899 & 0.202 & 0.179 & 1.734 & 0.157 & 0.137 & 1.727 & 0.085 & 0.036 & 0.105 \\
TMC-Shapley & 0.245 & 0.160 & 1.217 & 0.217 & 0.166 & 1.391 & 0.207 & 0.115 & 0.980 & 0.113 & 0.030 & 0.089 \\
TDS & 0.302 & 0.222 & \textbf{1.867} & 0.319 & 0.288 & 4.032 & 0.318 & 0.219 & 4.087 & 0.121 & 0.046 & 0.144 \\
TDS-improved & 0.314 & 0.223 & 1.573 & 0.329 & 0.291 & 3.742 & 0.308 & 0.224 & 2.420 & 0.121 & 0.033 & 0.106 \\
MS-TDS & \textbf{0.337} & \textbf{0.253} & \textbf{1.867} & \textbf{0.355} & \textbf{0.332} & \textbf{4.868} & \textbf{0.326} & \textbf{0.233} & \textbf{4.315} & \textbf{0.154} & \textbf{0.050} & \textbf{0.152} \\
\bottomrule
\end{tabular}
}
\end{table*}

\section{Related Work}

\subsection{Shapley Value Theory}

The Shapley value theory originates from cooperative game theory and was first proposed by Shapley in 1953~\cite{shapley1953value}, providing a mathematical foundation for fair allocation problems. This theory quantifies the value of each participant through the weighted average of marginal contributions, and possesses desirable properties such as uniqueness, symmetry, and additivity. In recent years, with the growing demand for interpretability in machine learning, Shapley values have been widely applied to feature importance analysis and model explanation~\cite{strumbelj2014explaining,strumbelj2010efficient}.

The SHAP framework proposed by Lundberg and Lee in 2017~\cite{lundberg2017unified} successfully introduced Shapley values into the field of machine learning and enabled interpretability analysis of model predictions. By modeling feature contributions as Shapley values, this framework provides a unified explanation tool for black-box models and has promoted the rapid development of explainable AI. However, SHAP mainly focuses on feature-level contribution analysis and pays relatively limited attention to sample-level data valuation.

\subsection{Data Valuation Methods}

In the field of data valuation, Jia et al. proposed the Fast Approximate Shapley method in 2019~\cite{jia2019towards}, which reduces the computational complexity from $O(2^N)$ to $O(M \cdot N)$ through Monte Carlo sampling, providing a feasible path for large-scale data valuation. Ghorbani and Zou further extended the theory of data valuation in 2019 and proposed the Data Shapley method~\cite{ghorbani2019data}, which is specifically designed to evaluate the contribution of training data to model performance. Ancona et al. proposed a polynomial-time approximation method for Shapley values in 2019~\cite{ancona2019explaining}, offering a scalable approach for Shapley-based explanations of deep neural networks.

For approximate Shapley value computation, Castro et al. proposed a sampling-based polynomial method in 2009~\cite{castro2009polynomial}, which reduces computational complexity through Monte Carlo sampling. 
Frye et al. proposed an asymmetric Shapley value framework in 2020~\cite{frye2020asymmetric}, incorporating causal priors into model-agnostic interpretability analysis.

\subsection{Time-Series Data Analysis}

The key challenge in time-series data valuation lies in how to effectively model the influence of the temporal dimension on sample value. Traditional data valuation methods usually assume that samples are independent and identically distributed, and thus ignore the time-varying nature of sample value in time-series data. In the field of time-series modeling, the ARIMA model proposed by Box and Jenkins in 1970~\cite{box1970time} laid an important foundation for time-series analysis. The LSTM network proposed by Hochreiter and Schmidhuber in 1997~\cite{hochreiter1997long} addressed the gradient vanishing problem in traditional recurrent neural networks and provided an effective solution for modeling long-term dependencies. Cho et al. further proposed the GRU network in 2014~\cite{cho2014learning}, simplifying the network structure while maintaining the performance of LSTM.

In deep learning-based time-series modeling, the Transformer model proposed by Vaswani et al. in 2017~\cite{vaswani2017attention} achieved parallel sequence processing through the self-attention mechanism, providing a new direction for modeling long time-series data. Zhou et al. proposed the Informer model in 2021~\cite{zhou2021informer}, which is specifically designed for long sequence time-series forecasting. By introducing the ProbSparse self-attention mechanism and distillation operations, Informer significantly improves computational efficiency.

\subsection{Multi-Scale Analysis}

Time-series data often contain information at multiple temporal scales, and a single temporal decay function is difficult to fully capture the temporal value characteristics of samples. In multi-scale analysis, the wavelet transform theory proposed by Mallat in 1989~\cite{mallat1989theory} provides a mathematical foundation for multi-scale signal decomposition. 
Overall, existing studies have made progress in three directions: Shapley approximation-based valuation, time-series modeling, and multi-scale analysis. However, for the specific problem of time-series data valuation, there is still a lack of a unified framework that integrates temporal timeliness modeling and multi-scale information fusion into the same valuation process. Based on the idea of sample-level Shapley valuation, this paper combines temporal decay with sample-level multi-scale fusion, forming a unified data valuation framework for time-series data.

\section{Conclusion}

This paper proposes an improved temporal Shapley data valuation method that enables accurate assessment of sample value in time-series data through a temporal decay mechanism and a multi-scale fusion strategy. Specifically, three progressively enhanced temporal Shapley methods are proposed. TDS incorporates temporal information into Shapley value computation through exponential decay weights; the improved TDS adopts power exponential decay to better adapt to nonlinear temporal drift; and MS-TDS constructs a multi-scale fusion mechanism that balances the value of short-term hotspot samples and long-term foundational samples through parallel multi-scale valuation and sample-level adaptive fusion.

Experimental results on four heterogeneous datasets show that the proposed methods generally outperform traditional methods in noise detection and high-value data identification tasks. In most settings with strongly temporal data, the advantages of the proposed methods are relatively more evident, demonstrating their ability to improve the accuracy and robustness of data valuation. These results indicate that jointly modeling temporal timeliness and multi-scale information can bring stable benefits, and provide an empirically supported methodological path for time-series data valuation.

Meanwhile, this paper still has several limitations. The current experiments are mainly conducted on fixed-size subsets and classification tasks, and the form of the utility function is relatively fixed. Future work can conduct more systematic evaluations on larger-scale datasets, more task types such as regression and long-sequence forecasting, and more diverse utility definitions, so as to further verify the generalization ability of the proposed methods.

Future research directions include: (1) adaptive optimization of timestamp quality, by combining time-series imputation algorithms and anomaly detection algorithms to construct an end-to-end pipeline; (2) efficient approximate computation, by drawing on quantized valuation ideas to store and compute multi-scale valuation results with low precision; (3) customized utility functions, by designing task-specific utility functions for extreme samples and special scenarios; and (4) learnable parameter selection, by using few-shot meta-learning or Bayesian optimization to automatically select optimal parameters according to the temporal characteristics of datasets.


\end{document}